\documentclass[10pt,twocolumn,letterpaper]{article}

\usepackage{cvpr}              

\usepackage{graphicx}
\usepackage{amsmath}
\usepackage{amssymb}
\usepackage{booktabs}

\usepackage[accsupp]{axessibility}

\usepackage{multirow}
\usepackage{bbding}

\usepackage[pagebackref,breaklinks,colorlinks]{hyperref}

\usepackage[capitalize]{cleveref}
\crefname{section}{Sec.}{Secs.}
\Crefname{section}{Section}{Sections}
\Crefname{table}{Table}{Tables}
\crefname{table}{Tab.}{Tabs.}

\def\cvprPaperID{****} 
\def\confName{CVPR}
\def\confYear{2023}

\begin{document}

\title{Language Adaptive Weight Generation for Multi-task Visual Grounding}

\author{
    Wei Su\textsuperscript{\rm 1}\quad
    Peihan Miao\textsuperscript{\rm 1}\quad
    Huanzhang Dou\textsuperscript{\rm 1}\quad
    Gaoang Wang\textsuperscript{\rm 4}\\
    Liang Qiao\textsuperscript{\rm 1,\rm 3}\quad
    Zheyang Li\textsuperscript{\rm 1,\rm 3}\quad
    Xi Li\textsuperscript{\rm 1,\rm 2,\rm 5\thanks{corresponding author.}}\\
    \textsuperscript{\rm 1}Zhejiang University\quad
    \textsuperscript{\rm 2}Shanghai AI Laboratory\quad
    \textsuperscript{\rm 3}Hikvision Research Institute\\
    \textsuperscript{\rm 4}Zhejiang University-University of Illinois Urbana-Champaign Institute, Zhejiang University\\
    \textsuperscript{\rm 5}Shanghai Institute for Advanced Study of Zhejiang University\\
    {\tt\small \{weisuzju, peihan.miao, hzdou, qiaoliang, xilizju\}@zju.edu.cn}\\
    {\tt\small gaoangwang@intl.zju.edu.cn}, 
    {\tt\small lizheyang@hikvision.com}
}

\maketitle

\begin{abstract}
Although the impressive performance in visual grounding, the prevailing approaches usually exploit the visual backbone in a passive way, \textit{i.e.,} the visual backbone extracts features with fixed weights without expression-related hints.
The passive perception may lead to mismatches (\textit{e.g.,} redundant and missing), limiting further performance improvement.
Ideally, the visual backbone should actively extract visual features since the expressions already provide the blueprint of desired visual features.
The active perception can take expressions as priors to extract relevant visual features, which can effectively alleviate the mismatches.
Inspired by this, we propose an active perception \textbf{V}isual \textbf{G}rounding framework based on \textbf{L}anguage \textbf{A}daptive \textbf{W}eights, called VG-LAW.
The visual backbone serves as an expression-specific feature extractor through dynamic weights generated for various expressions.
Benefiting from the specific and relevant visual features extracted from the language-aware visual backbone, VG-LAW does not require additional modules for cross-modal interaction.
Along with a neat multi-task head, VG-LAW can be competent in referring expression comprehension and segmentation jointly.
Extensive experiments on four representative datasets, \textit{i.e.,} RefCOCO, RefCOCO+, RefCOCOg, and ReferItGame, validate the effectiveness of the proposed framework and demonstrate state-of-the-art performance.
\end{abstract}
\section{Introduction}
Visual grounding (such as referring expression comprehension \cite{mattnet,resc,transvg,qrnet,word2pix,seqtr,rt}, referring expression segmentation \cite{mcn,lavt,rt,vlt,lts,cgan,hu2016segmentation}, and phrase grounding \cite{seqtr,transvg,rt}) aims to detect or segment the specific object based on a given natural language description.
Compared to general object detection \cite{faster} or instance segmentation \cite{he2017mask}, which can only locate objects within a predefined and fixed category set, visual grounding is more flexible and purposeful. 
Free-formed language descriptions can specify specific visual properties of the target object, such as categories, attributes, relationships with other objects, relative/absolute positions, and \etc.

\begin{figure}
    \centering
    \includegraphics[width=0.48\textwidth]{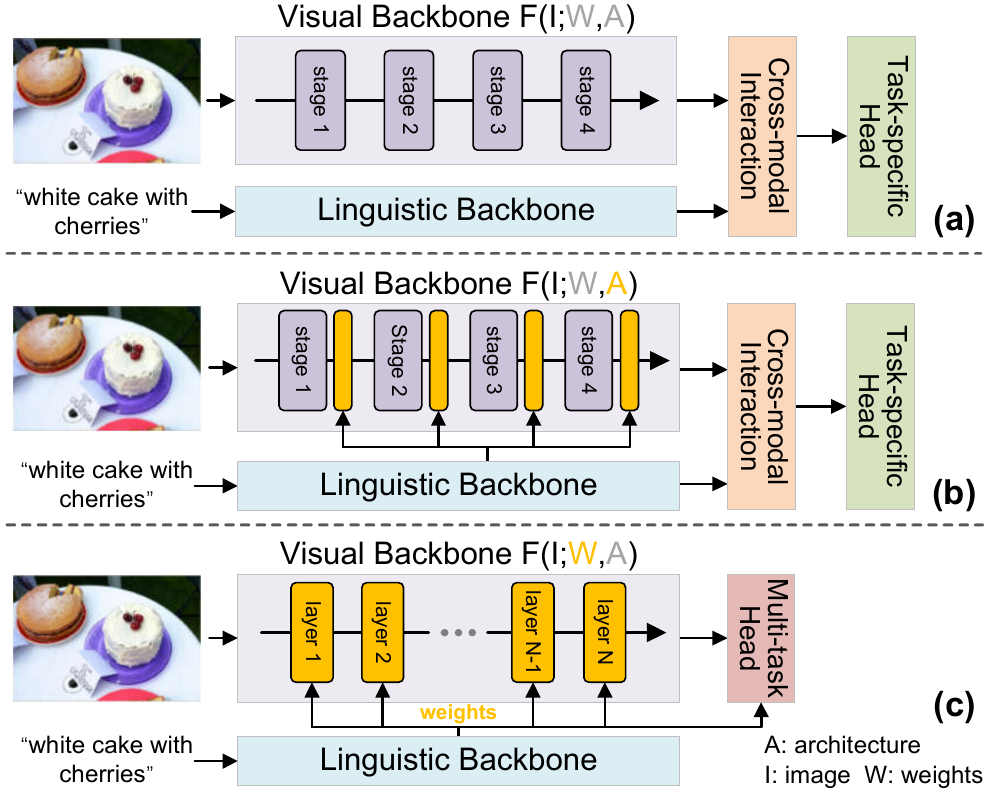}
    \caption{The comparison of visual grounding frameworks. (a) The visual and linguistic backbone independently extracts features, which are fused through cross-modal interaction. (b) Additional designed modules are inserted into the visual backbone to modulate visual features using linguistic features. (c) VG-LAW can generate language-adaptive weights for the visual backbone and directly output referred objects through our designed multi-task head without additional cross-modal interaction modules.}
    \label{fig:first_fig}
\end{figure}

\begin{figure}
    \centering
    \includegraphics[width=0.47\textwidth]{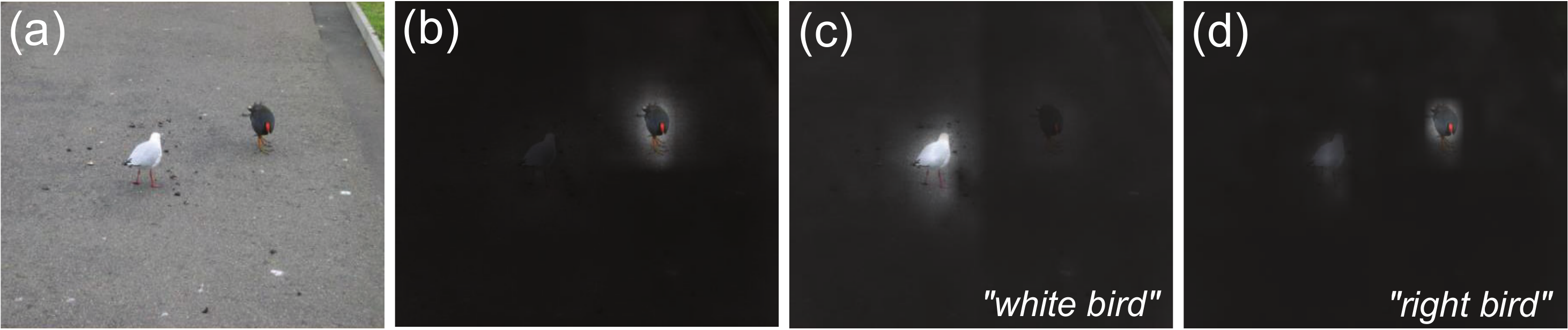}
    \caption{Attention visualization of the visual backbone with different weights. (a) input image, (b) visual backbone with fixed weights, (c) and (d) visual backbone with weights generated for ``white bird" and ``right bird", respectively.}
    \label{fig:second_fig}
\end{figure}

Due to the similarity with detection tasks, previous visual grounding approaches \cite{mattnet,mcn,seqtr,rt} usually follow the general object detection frameworks \cite{he2017mask,redmon2018yolov3,detr}, and pay attention to the design of cross-modal interaction modules.
Despite achieving impressive performance, the visual backbone is not well explored.
Concretely, the visual backbone passively extracts visual features with fixed architecture and weights, regardless of the referring expressions, as illustrated in \cref{fig:first_fig} (a).
Such passive feature extraction may lead to mismatches between the extracted visual features and those required for various referring expressions, such as missing or redundant features.
Taking \cref{fig:second_fig} as an example, the fixed visual backbone has an inherent preference for the image, as shown in \cref{fig:second_fig} (b), which may be irrelevant to the referring expression ``white bird".
Ideally, the visual backbone should take full advantage of expressions, as the expressions can provide information and tendencies about the desired visual features.


Several methods have noticed this phenomenon and proposed corresponding solutions, such as QRNet \cite{qrnet}, and LAVT \cite{lavt}. 
Both methods achieve the expression-aware visual feature extraction by inserting carefully designed interaction modules (such as QD-ATT \cite{qrnet}, and PWAN \cite{lavt}) into the visual backbone, as illustrated in \cref{fig:first_fig} (b).
Concretely, visual features are first extracted and then adjusted using QD-ATT (channel and spatial attention) or PWAM (transformer-based pixel-word attention) in QRNet and LAVT at the end of each stage, respectively.
Although performance improvement with adjusted visual features, the extract-then-adjust paradigm inevitably contains a large number of feature-extraction components with fixed weights, \textit{e.g.,} the components belonging to the original visual backbone in QRNet and LAVT.
Considering that the architecture and weights jointly determine the function of the visual backbone, this paper adopts a simpler and fine-grained scheme that modifies the function of the visual backbone with language-adaptive weights, as illustrated in \cref{fig:first_fig} (c).
Different from the extract-then-adjust paradigm used by QRNet and LAVT, the visual backbone equipped with language-adaptive weights can directly extract expression-relevant visual features without additional feature-adjustment modules.

In this paper, we propose an active perception \textbf{V}isual \textbf{G}rounding framework based on \textbf{L}anguage \textbf{A}daptive \textbf{W}eights, called VG-LAW.
It can dynamically adjust the behavior of the visual backbone by injecting the information of referring expressions into the weights.
Specifically, VG-LAW first obtains the specific language-adaptive weights for the visual backbone through two successive processes of linguistic feature aggregation and weight generation.
Then, the language-aware visual backbone can extract expression-relevant visual features without manually modifying the visual backbone architecture.
Since the extracted visual features are highly expression-relevant, cross-modal interaction modules are not required for further cross-modal fusion, and the entire network architecture is more streamlined.
Furthermore, based on the expression-relevant features, we propose a lightweight but neat multi-task prediction head for jointly referring expression comprehension (REC) and referring expression segmentation (RES) tasks.
Extensive experiments on RefCOCO \cite{refcoco}, RefCOCO+ \cite{refcoco}, RefCOCOg \cite{refcocog-umd}, and ReferItGame \cite{referitgame} datasets demonstrate the effectiveness of our method, which achieves state-of-the-art performance.

The main contributions can be summarized as follows:
\begin{itemize}
    \item We propose an active perception visual grounding framework based on the language adaptive weights, called VG-LAW, which can actively extract expression-relevant visual features without manually modifying the visual backbone architecture.
    \item Benefiting from the active perception of visual feature extraction, we can directly utilize our proposed neat but efficient multi-task head for REC and RES tasks jointly without carefully designed cross-modal interaction modules. 
    \item Extensive experiments demonstrate the effectiveness of our framework, which achieves state-of-the-art performance on four widely used datasets, \textit{i.e.,} RefCOCO, RefCOCO+, RefCOCOg, and ReferItGame.
\end{itemize}

\begin{figure*}[h]
    \centering
    \includegraphics[width=\textwidth]{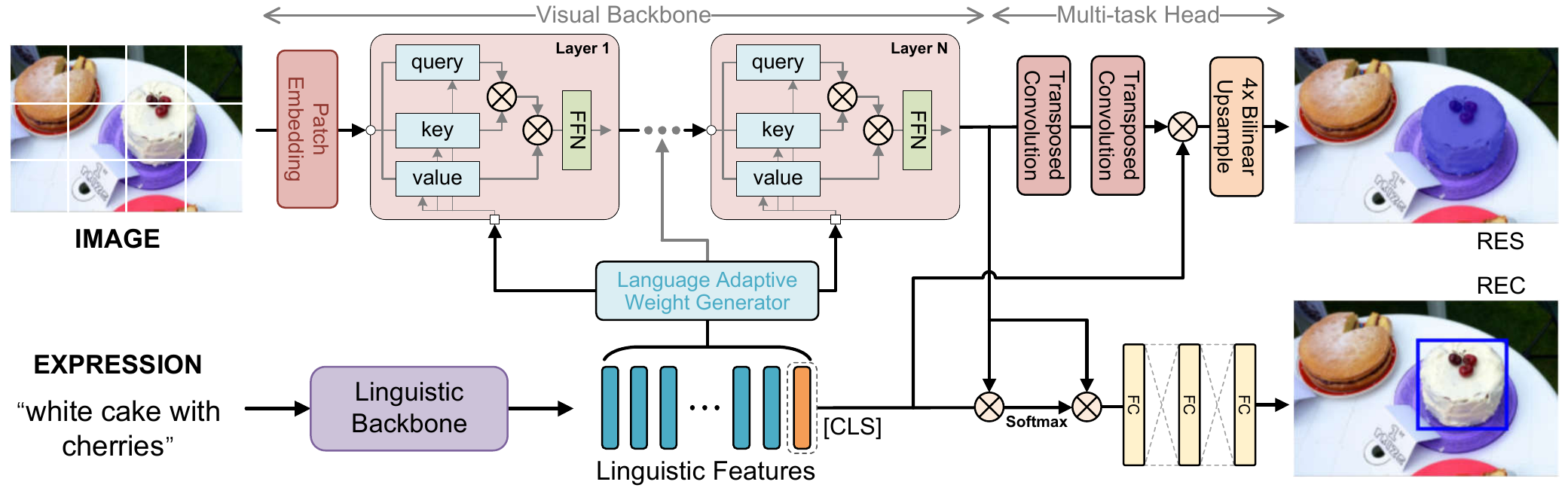}
    \caption{The overall architecture of our proposed VG-LAW framework. It consists of four components: (1) Linguistic Backbone, which extracts linguistic features from free-formed referring expressions, (2) Language Adaptive Weight Generator, which generates dynamic weights for the visual backbone conditioned on specific expressions, (3) Visual Backbone, which extracts visual features from the raw image and its behavior can be modified by language-adaptive weights, and (4) Multi-task Head, which predicts the bounding box and mask of referred object jointly. 
   $\otimes$ represents the matrix multiplication.}
    \label{fig:main_arch}
\end{figure*}

\section{Related Work}

\subsection{Referring Expression Comprehension}
Referring expression comprehension (REC) \cite{mattnet,cm-att-erase,rvg-tree,resc,yang2019fast,realgin,transvg,word2pix,seqtr} aims to generate a bounding box in an image specified by a given referring expression.
Early researchers explore REC through a two-stage framework \cite{mattnet,cm-att-erase,liu2019learning,rvg-tree}, where region proposals \cite{faster} are first extracted and then ranked according to their similarity scores with referring expressions.
To alleviate the speed and accuracy issues of the region proposals in the two-stage framework, simpler and faster one-stage methods \cite{resc,yang2019fast,realgin} based on dense anchors are proposed.
Recently, transformer-based methods \cite{transvg,word2pix,seqtr,mdetr,yoro} can effectively capture intra- and inter-modality context and achieve better performance, benefiting from the self-attention mechanism\cite{transformer}.

\subsection{Referring Expression Segmentation}
Similar to REC, referring expression segmentation (RES) \cite{hu2016segmentation,efn,cgan,lts,cmpc,restr,lavt,rt,seqtr,vlt} aims to predict a precise pixel-wise binary mask corresponding to the given referring expression.
The pioneering work \cite{hu2016segmentation} proposes to generate segmentation masks 
for natural language expressions by concatenating the visual and linguistic features and mixing these two modal features with fully convolutional classifiers.
Follow-up solutions \cite{efn,cgan,lts,cmpc} propose various attention mechanisms to perform cross-modal interaction to generate a high-resolution segmentation map.
Recent studies \cite{restr,lavt,rt,seqtr,vlt}, like REC, leverage transformer \cite{transformer} to realize cross-modal interaction and achieve excellent performance. All these methods achieve cross-modal interaction by either adjusting the inputs or modifying the architectures with fixed network weights. 

\subsection{Dynamic Weight Networks}
Several works \cite{ha2016hypernetworks,jia2016dynamicfilternet,chen2020dynamicconv,yang2019condconv,li2020dcd} have investigated dynamic weight networks, where given inputs adaptively generate the weights of the network.
According to the way of dynamic weight generation, the current methods can be roughly divided into three categories. 
(1) Dynamic weights are directly generated using fully-connected layers with learnable embeddings \cite{ha2016hypernetworks} or intermediate features \cite{jia2016dynamicfilternet} as input.
(2) Weights are computed as the weighted sum of a set of learnable weights \cite{chen2020dynamicconv, yang2019condconv,li2022omni}, which can also be regarded as the mixture-of-experts and may suffer from challenging joint optimization.
(3) The weights are analyzed from the perspective of matrix decomposition \cite{li2020dcd}, and the final dynamic weights are generated by calculating the multiplication of several matrices.
\section{Method} 
In this section, we will introduce the active perception framework for multi-task visual grounding, including the language-adaptive weight generation, multi-task prediction head, and training objectives.

\subsection{Overview}
\label{method:sec1}
The extraction of visual features by the visual backbone in the manner of passive perception may cause mismatch problems, which can lead to suboptimal performance despite subsequent carefully designed cross-modal interaction modules.
Considering that expressions already provide a blueprint for the desired visual features, we propose an active perception visual grounding framework based on the language adaptive weights, called VG-LAW, as illustrated in \cref{fig:main_arch}.
In this framework, the visual backbone can actively extract expression-relevant visual features using language-adaptive weights, without needing to manually modify the visual backbone architecture or elaborately design additional cross-modal interaction modules.

Specifically, the VG-LAW framework consists of four components, \textit{i.e.,} linguistic backbone, language adaptive weight generator, visual backbone, and multi-task head.
Given a referring expression, the $N$-layer BERT-based \cite{bert} linguistic backbone tokenizes the expressions, prepends a [CLS] token, and extracts linguistic features $F_l \in \mathbb{R}^{L \times d_l}$, where $L$ and $d_l$ represent the token numbers and dimension of linguistic features, respectively.
The linguistic features $F_l$ are then fed to the language adaptive weight generator to generate weights for the transformer-based visual backbone.
Next, given an image $I \in \mathbb{R}^{3 \times H \times W}$, the expression-aware visual features $F_v \in \mathbb{R}^{C \times \frac{H}{s} \times \frac{W}{s}}$ can be extracted by the visual backbone, where $C$ and $s$ represent the channel number and stride of the visual features, respectively.
Finally, we pass the linguistic features $F_l^1 \in \mathbb{R}^{d_l}$ represented by the [CLS] token and the visual features to the multi-task head, which predicts the bounding box and mask of the referred object for REC and RES, respectively.

\begin{figure}
    \centering
    \includegraphics[width=0.47\textwidth]{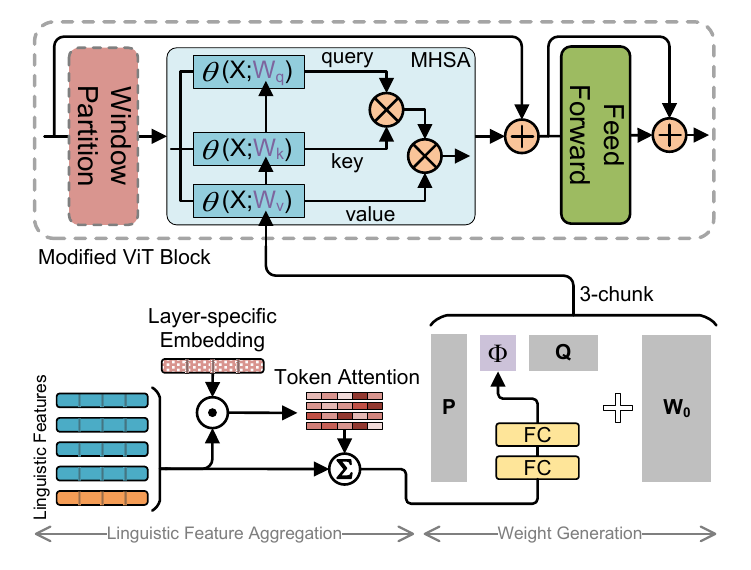}
    \caption{The detailed architecture for language adaptive weight generation. The upper part shows the architecture of the adapted ViT block in the visual backbone, and the lower part shows the linguistic feature aggregation and weight generation.}
    \label{fig:weight_generator}
\end{figure}

\subsection{Language Adaptive Weight Generation}
\label{method:sec2}
After extracting linguistic features, language-adaptive weights are generated to guide the active perception of the visual backbone.
The process of language adaptive weight generation has two stages, \textit{i.e.,} the layer-wise linguistic feature aggregation and the weight generation.

\paragraph{Linguistic Feature Aggregation.}
Considering the referring expressions correspond to a different number of linguistic tokens and each layer of the visual backbone may prefer different linguistic tokens, we try to aggregate linguistic features with fixed sizes for each layer independently.
Inspired by the multi-head attention mechanism \cite{transformer}, we introduce a learnable layer-specific embedding $e_i \in \mathbb{R}^{d_l}$ for each layer $i$ of the visual backbone to extract layer-specific linguistic features dynamically, which can improve the model flexibility at negligible cost. The calculation is performed on $G$ groups. For each group $g$, the token-wise attention $\alpha_i^g \in [0,1]^{L}$ is assigned to the normalized dot product of $e_i^g$ and $F_l^g$, which is denoted as:  
\begin{equation}
    \alpha_i^g = \mathrm{Softmax}([e_i^g\cdot F_l^{g,1}, e_i^g\cdot F_l^{g,2}, \cdots, e_i^g\cdot F_l^{g,L}]).
    \label{equ:token_attn}
\end{equation}
Then, the aggregated linguistic feature $h_0^i \in \mathbb{R}^{d_l}$ can be derived by concatenating $h_0^{i,g}=\sum_{j=1}^{L}\alpha_i^{g,j} F_l^{g,j}$.

Finally, we use a fully-connected layer (FC) to reduce the dimension of the aggregated linguistic features for the $i$-th layer of the visual backbone, which is indicated as:
\begin{equation}
    h_{1}^{i} = \delta(W_1^i h_{0}^{i}),
    \label{equ:inner_feat_reduce}
\end{equation}
where $W_1^i \in \mathbb{R}^{d_l \times d_h}$ is used to reducing the dimension to $d_h=d_l/r$, and $r$ is the reduction ratio. $\delta$ refers to the $\mathrm{GeLU}$ activation function.

\paragraph{Weight Generation.}
\label{method:sec2.2}
To guide the active perception of the visual backbone, we generate language-adaptive weights for producing the query $X_q$, key $X_k$, and value $X_v$ in the visual backbone conditioned on referring expressions, which can be represented as:
\begin{equation}
    X_q=\theta(X;W_q),X_k=\theta(X;W_k),X_v=\theta(X;W_v),
    \label{equ:qkv}
\end{equation}
where $\theta(\cdot;W)$ indicates the linear projection operation parameterized by $W$, and $X$ represents the input visual features. $W_q,W_k,W_v$$\in$$\mathbb{R}^{d_{out}\times d_{in}}$ are the dynamic projection weights used to generate the query, key, and value, respectively. $d_{in}$ and $d_{out}$ are the dimension of feature $X$ and query/key/value, respectively.

Considering the large number $d_{out}\times d_{in}$ of the dynamic weights, it is unaffordable to directly generate weights using fully-connected layers like Hypernetworks \cite{ha2016hypernetworks}. 
The DynamicConv \cite{chen2020dynamicconv} and CondConv \cite{yang2019condconv} can alleviate this problem by generating weights with weighted summation of $K$ static kernels but can increase the parameter number by $K$-times and suffer from challenging joint optimization.
Inspired by the dynamic channel fusion \cite{li2020dcd}, we try to generate dynamic weights following the matrix decomposition paradigm. 
Taking the $i$-th ViT block as an example, which can be formulated as:
\begin{equation}
    [W_q^i,W_k^i,W_v^i] = W_0^i + P\Phi(h_{1}^{i})Q^T,
    \label{equ:weight_qkv}
\end{equation}
where $W_0^i\in \mathbb{R}^{d_{out}\times d_{in}}$ is the layer-specific static learnable weights. $P\in \mathbb{R}^{d_{out}\times d_{w}}$ and $Q\in \mathbb{R}^{d_{in}\times d_{w}}$ are also static learnable weights, but sharable across all ViT blocks to reduce the parameter numbers and prevent the model from overfitting. $\Phi(h_{1}^{i})$ is a fully-connected layer, which produces a dynamic matrix of shape $d_w \times d_w$ with aggregated linguistic features $h_1^i$ as input.

\subsection{Multi-task Head}
\label{method:sec3}
Different from the previous methods \cite{resc,qrnet,seqtr,rt,mattnet,realgin,vlt}, which require carefully designed cross-modal interaction modules, VG-LAW can obtain expression-relevant visual features extracted by the language-aware visual backbone without additional cross-modal interaction modules.
Through our proposed neat but efficient multi-task head, we can utilize the visual and linguistic features to predict the bounding box for REC and the segmentation mask for RES.
Concretely, there are two branches in the multi-task head for REC and RES, respectively.

For the REC branch, we apply direct coordinate regression to predict the bounding box of referred object.
To pool the 2-$d$ visual features along the spatial dimension, we propose a language adaptive pooling module (LAP), which aggregates visual features using language-adaptive attention.
Specifically, the visual features $\{F_v^{i,j}\}\in \mathbb{R}^{C \times \frac{H}{s} \times \frac{W}{s}}$ and linguistic feature $F_l^1 \in \mathbb{R}^{d_l}$ are firstly projected to the lower-dimension space $\mathbb{R}^k$, and the attention weights $A \in \mathbb{R}^{\frac{H}{s} \times \frac{W}{s}}$ are calculated as dot-product similarity followed by $\mathrm{Softmax}$ normalization.
Then, the visual features are aggregated by calculating the weighted sum with attention weights $A$.
Finally, the aggregated visual features are fed to a three-layer fully-connected layer, and the Sigmoid function is used to predict the referred bounding box $\hat{b}=(\hat{x},\hat{y},\hat{w},\hat{h})$.

For the RES branch, we apply binary classification to each visual feature along the spatial dimension to predict segmentation masks for referred objects.
Specifically, the visual features $F_v$ are first up-sampled to $\hat{F}_v \in \mathbb{R}^{d_l \times \frac{H}{4} \times \frac{W}{4}}$ with successive transposed convolutions.
Then, the intermediate segmentation map $\bar{s} \in \mathbb{R}^{\frac{H}{4} \times \frac{W}{4}}$ can be obtained by using linear projection $\theta(\cdot;W)$ on each visual feature.
Following the language adaptive weight paradigm, we also use dynamic rather than fixed weights by simply setting $W=F_l^1$.
Finally, the full-resolution segmentation mask $\hat{s} \in \mathbb{R}^{H \times W}$ is derived by simply up-sample $\bar{s}$ using bilinear interpolation, followed by the Sigmoid function.

\subsection{Training Objectives}
\label{method:sec4}
The VG-LAW framework can be optimized end-to-end for multi-task visual grounding.
For REC, given the predicted bounding box $\hat{b}=(\hat{x},\hat{y},\hat{w},\hat{h})$ and the ground truth $b=(x,y,w,h)$, the detection loss function is defined as follows:
\begin{equation}
    \mathcal{L}_{det} = \lambda_{L1}\mathcal{L}_{L1}(b,\hat{b}) + \lambda_{giou}\mathcal{L}_{giou}(b,\hat{b}),
\end{equation}
where $\mathcal{L}_{L1}(\cdot,\cdot)$ and $\mathcal{L}_{giou}(\cdot,\cdot)$ represent L1 loss and Generalized IoU loss \cite{giou}, respectively, and $\lambda_{L1}$ and $\lambda_{giou}$ are the relative weights to control the two detection loss functions.
For RES, given the predicted mask $\hat{s}$ and the ground-truth $s$, the segmentation loss function is defined as follows:
\begin{equation}
    \mathcal{L}_{seg} = \lambda_{focal}\mathcal{L}_{focal}(s,\hat{s}) + \lambda_{dice}\mathcal{L}_{dice}(s,\hat{s}),
\end{equation}
where $\mathcal{L}_{focal}(\cdot,\cdot)$ and $\mathcal{L}_{dice}(\cdot,\cdot)$ represent focal loss \cite{focal} and DICE/F-1 loss \cite{dice}, respectively, and $\lambda_{focal}$ and $\lambda_{dice}$ are the relative weights to control the two segmentation loss functions. 
Our framework can be seamlessly used for joint training of REC and RES, and its joint training loss function is defined as follows:
\begin{equation}
    \mathcal{L}_{total} = \mathcal{L}_{det} + \mathcal{L}_{seg}.
\end{equation}

The trained model performs well for language-guided detection and segmentation. 
The experimental analysis of the whole framework will be elaborated in \cref{experiments}.
\begin{table*}[h]
    \footnotesize
    \centering
      \begin{tabular}{c|c|c|c|ccc|ccc|cc|c}
      \toprule
      \multirow{2}[2]{*}{Methods} & \multirow{2}[2]{*}{Venue} & Visual & Multi- & \multicolumn{3}{c|}{RefCOCO} & \multicolumn{3}{c|}{RefCOCO+} & \multicolumn{2}{c|}{RefCOCOg} & ReferItGame \\
            &       & Backbone & task  & val   & testA & testB & val   & testA & testB & val   & test  & test \\
      \midrule
      \textbf{Two-stage:} &       &       &       &       &       &       &       &       &       &       &       &  \\
      MAttNet \cite{mattnet} &   CVPR18   & RN101 &   \XSolidBrush    & 76.65  & 81.14  & 69.99  & 65.33  & 71.62  & 56.02  & 66.58  & 67.27  & 29.04  \\
      RvG-Tree \cite{rvg-tree} &   TPAMI19   & RN101 &   \XSolidBrush    & 75.06  & 78.61  & 69.85  & 63.51  & 67.45  & 56.66  & 66.95  & 66.51  & - \\
      CM-A-E  \cite{cm-att-erase} &   CVPR19   & RN101 &   \XSolidBrush    & 78.35  & 83.14  & 71.32  & 68.09  & 73.65  & 58.03  & 67.99  & 68.67  & - \\
      Ref-NMS \cite{ref-nms} &   AAAI21  & RN101 &   \XSolidBrush  & 80.70  & 84.00  & 76.04  & 68.25  & 73.68  & 59.42  & 70.55  & 70.62  & - \\
      \midrule
      \midrule
      \textbf{One-stage:} &       &       &       &       &       &       &       &       &       &       &       &  \\
      FAOA \cite{yang2019fast} &   ICCV19  & DN53  &   \XSolidBrush   & 72.54  & 74.35  & 68.50  & 56.81  & 60.23  & 49.60  & 61.33  & 60.36  & 60.67  \\
      ReSC-Large \cite{resc} &    ECCV20   & DN53  &   \XSolidBrush    & 77.63  & 80.45  & 72.30  & 63.59  & 68.36  & 56.81  & 67.30  & 67.20  & 64.60  \\
      MCN \cite{mcn}  &  CVPR20  & DN53  &  \Checkmark  & 80.08  & 82.29  & 74.98  & 67.16  & 72.86  & 57.31  & 66.46  & 66.01  & - \\
      RealGIN \cite{realgin} &   TNNLS21   & DN53  &    \XSolidBrush    & 77.25  & 78.70  & 72.10  & 62.78  & 67.17  & 54.21  & 62.75  & 62.33  & - \\
      PLV-FPN* \cite{plv} &   TIP22    & RN101 &   \XSolidBrush    & 81.93  & 84.99  & 76.25  & 71.20  & 77.40  & 61.08  & 70.45  & 71.08  & 71.77  \\
      \midrule
      \midrule
      \textbf{Transformer-based:} &       &       &       &       &       &       &       &       &       &       &       &  \\
      TransVG \cite{transvg} &   ICCV21    & RN101 & \XSolidBrush
      & 81.02  & 82.72  & 78.35  & 64.82  & 70.70  & 56.94  & 68.67  & 67.73  & 70.73  \\
      RefTR* \cite{rt} & NeurIPS21 & RN101 & \Checkmark  & 82.23  & 85.59  & 76.57  & 71.58  & 75.96  & 62.16  & 69.41  & 69.40  & 71.42  \\
      SeqTR \cite{seqtr} & ECCV22  & DN53  & \XSolidBrush & 81.23  & 85.00  & 76.08  & 68.82  & 75.37  & 58.78  & 71.35  & 71.58  & 69.66  \\
      Word2Pix \cite{word2pix} &  TNNLS22  & RN101 &  \XSolidBrush  & 81.20  & 84.39  & 78.12  & 69.74  & 76.11  & 61.24  & 70.81  & 71.34  & - \\
      YORO \cite{yoro} &  ECCVW22  & - &  \XSolidBrush  & 82.90  & 85.60  & 77.40  & 73.50  & 78.60  & 64.90  & 73.40  & 74.30  & 71.90 \\
      QRNet \cite{qrnet} & CVPR22 & Swin-S & \XSolidBrush  & 84.01  & 85.85  & 82.34  & 72.94  & 76.17  & 63.81  & 71.89  & 73.03  & 74.61  \\
      \midrule
      \midrule
      \textbf{Ours:} &       &       &       &       &       &       &       &       &       &       &       &  \\
        VG-LAW    &   -    &   ViT-B    &   \XSolidBrush    &    \textcolor{blue}{86.06}   &   \textcolor{blue}{88.56}    &    \textcolor{blue}{82.87}   &    \textcolor{blue}{75.74}   &    \textcolor{blue}{80.32}   &     \textcolor{blue}{66.69}  &    \textcolor{blue}{75.31}   &     \textcolor{blue}{75.95}  & \textcolor{blue}{76.60} \\
        VG-LAW    &   -   &   ViT-B    &   \Checkmark    &   \textcolor{red}{86.62}    &   \textcolor{red}{89.32}    &   \textcolor{red}{83.16}    &   \textcolor{red}{76.37}    &   \textcolor{red}{81.04}    &   \textcolor{red}{67.50}    &   \textcolor{red}{76.90}    &   \textcolor{red}{76.96}    &  \textcolor{red}{77.22} \\
      \bottomrule
      \end{tabular}%
      \caption{Comparison with state-of-the-art methods on RefCOCO \cite{refcoco}, RefCOCO+ \cite{refcoco}, RefCOCOg \cite{refcocog-umd} and ReferItGame \cite{referitgame} for REC task. The visual backbone is pre-trained on MS-COCO \cite{mscoco}, where overlapping images of the val/test sets are excluded. * represents ImageNet \cite{imagenet} pre-training. RN101, DN53, Swin-S, and ViT-B are shorthand for the ResNet101, DarkNet53, Swin-Transformer Small, and ViT Base, respectively. We highlight the best and second best performance in the \textcolor{red}{red} and \textcolor{blue}{blue} colors. }
    \label{tab:rec_sota}%
  \end{table*}%
  
\begin{table*}[htbp]
    \footnotesize
    \centering
      \begin{tabular}{c|c|c|c|ccc|ccc|cc}
      \toprule
      \multirow{2}[2]{*}{Methods} & \multirow{2}[2]{*}{Venue} & Visual & Multi- & \multicolumn{3}{c|}{RefCOCO} & \multicolumn{3}{c|}{RefCOCO+} & \multicolumn{2}{c}{RefCOCOg} \\
            &       & Backbone & task  & val   & testA & testB & val   & testA & testB & val   & test \\
      \midrule
      CGAN \cite{cgan} & MM20  & DN53  &   \XSolidBrush    & 64.86  & 68.04  & 62.07  & 51.03  & 55.51  & 44.06  & 54.40  & 54.25  \\
      MCN \cite{mcn}   & CVPR20 & DN53  &   \Checkmark    & 62.44  & 64.20  & 59.71  & 50.62  & 54.99  & 44.69  & 49.22  & 49.40  \\
      LTS \cite{lts}  & CVPR21 & DN53  &   \XSolidBrush    & 65.43  & 67.76  & 63.08  & 54.21  & 58.32  & 48.02  & 54.40  & 54.25  \\
      VLT \cite{seqtr}  & ICCV21 & DN53  &   \XSolidBrush    & 65.65  & 68.29  & 62.73  & 55.50  & 59.20  & 49.36  & 52.99  & 56.65  \\
      RefTR* \cite{rt} & NeurIPS21 & RN101 &   \Checkmark    & 70.56  & 73.49  & 66.57  & 61.08  & 64.69  & 52.73  & 58.73  & 58.51  \\
      SeqTR \cite{seqtr} & ECCV22 & DN53  &   \XSolidBrush    & 67.26  & 69.79  & 64.12  & 54.14  & 58.93  & 48.19  & 55.67  & 55.64  \\
      LAVT* \cite{lavt} & CVPR22 & Swin-B &   \XSolidBrush    & 74.46  & 76.89  & 70.94  & 65.81  & \textcolor{red}{70.97}  & \textcolor{red}{59.23}  & 63.62  & 63.66  \\
      \midrule
      \textbf{Ours:} &       &       &       &       &       &       &       &       &       &       &  \\
      VG-LAW & -     & ViT-B &   \XSolidBrush    &   \textcolor{blue}{75.05}    &    \textcolor{blue}{77.36}   &    \textcolor{blue}{71.69}   & \textcolor{blue}{66.61} & 70.30  &    58.14   &     \textcolor{blue}{65.36}  & \textcolor{blue}{65.13} \\
      VG-LAW & -     & ViT-B &   \Checkmark    &    \textcolor{red}{75.62}    &    \textcolor{red}{77.51}    &    \textcolor{red}{72.89}    &    \textcolor{red}{66.63}    &    \textcolor{blue}{70.38}    &    \textcolor{blue}{58.89}    &    \textcolor{red}{65.63}    &  \textcolor{red}{66.08} \\
      \bottomrule
      \end{tabular}%
      \caption{Comparison with state-of-the-art methods on RefCOCO \cite{refcoco}, RefCOCO+ \cite{refcoco}, and RefCOCOg \cite{refcocog-umd} for RES task. The visual backbone is pre-trained on MS-COCO \cite{mscoco}, where overlapping images of the val/test sets are excluded. * represents ImageNet \cite{imagenet} pre-training. RN101, DN53, Swin-B, and ViT-B are shorthand for the ResNet101, DarkNet53, Swin-Transformer Base, and ViT Base, respectively. We highlight the best and second best performance in the \textcolor{red}{red} and \textcolor{blue}{blue} colors.}
    \label{tab:res_sota}%
  \end{table*}%

\section{Experiments}
\label{experiments}

In this section, we will give a detailed experimental analysis of the whole framework, including the datasets, evaluation protocol, implementation details, comparisons with the state-of-the-art methods, and ablation analysis.

\subsection{Datasets and Evaluation Protocol}
\label{experiments:sec1}
\paragraph{Datasets.}
To verify the effectiveness of our method, we conduct experiments on the widely used RefCOCO \cite{refcoco}, RefCOCO+ \cite{refcoco}, RefCOCOg \cite{refcocog}, and ReferItGame \cite{referitgame} datasets.
RefCOCO, RefCOCO+, and RefCOCOg are collected from MS-COCO \cite{mscoco}.
RefCOCO and RefCOCO+, which are collected in interactive games, can be divided into train, val, testA, and testB sets.
Compared to RefCOCO, the expressions of RefCOCO+ contain more attributes than absolute locations.
Unlike RefCOCO and RefCOCO+, RefCOCOg collected by Amazon Mechanical Turk has a longer length of 8.4 words, including the attribute and location of referents.
Following a common version of split \cite{refcocog-umd}, RefCOCOg has train, val, and test sets.
In addition, ReferItGame collected from SAIAPR-12 \cite{saiapr12} contains train and test sets.
Each sample in the above datasets contains its corresponding bounding box and mask.

\paragraph{Evaluation Protocol.}
Following the previous works \cite{seqtr,mcn,rt}, we use \emph{Prec@0.5} and \emph{mIoU} to evaluate the performance of REC and RES, respectively.
For \emph{Prec@0.5}, the predicted bounding box is considered correct if the intersection-over-union (IoU) with the ground-truth bounding box is greater than 0.5.
\emph{mIoU} represents the IoU between the prediction and ground truth averaged across all test samples.

\subsection{Implementation Details}
\label{experiments:sec2}
\paragraph{Training.}
The resolution of the input image is resized to $448 \times 448$. 
ViT-Base \cite{vit} is used as the visual backbone, and we follow the adaptation introduced by ViTDet \cite{li2022vitdet} to adapt the visual backbone to higher-resolution images.
The visual backbone is pre-trained using Mask R-CNN \cite{he2017mask} on MS-COCO \cite{mscoco}, where overlapping images of the val/test sets are excluded.
The $W_0^i$ and $\Phi(h_1^i)$ in \cref{equ:weight_qkv} are initialized with the corresponding pre-trained weights of the visual backbone and zeros, respectively.
The maximum length of referring expression is set to 40, and the uncased base of six-layer BERT \cite{bert} as the linguistic backbone is used to generate linguistic features.
$\lambda_{L1}$ and $\lambda_{giou}$ are set to 1. $\lambda_{focal}$ and $\lambda_{dice}$ are set to 4.
The reduction ratio $r$ is set to 16.
The initial learning rate for the visual and linguistic backbone is 4e-5, and the initial learning rate for the remaining components is 4e-4.
The model is end-to-end optimized by AdamW \cite{adamw} for 90 epochs with a batch size of 256, where weight decay is set to 1e-4, and the learning rate is reduced by a factor of 10 after 60 epochs.
Data augmentation operation includes random horizontal flips.
We implement our framework using PyTorch and conduct experiments with NVIDIA A100 GPUs.

\begin{figure}[h]
    \centering
    \includegraphics[width=0.43\textwidth]{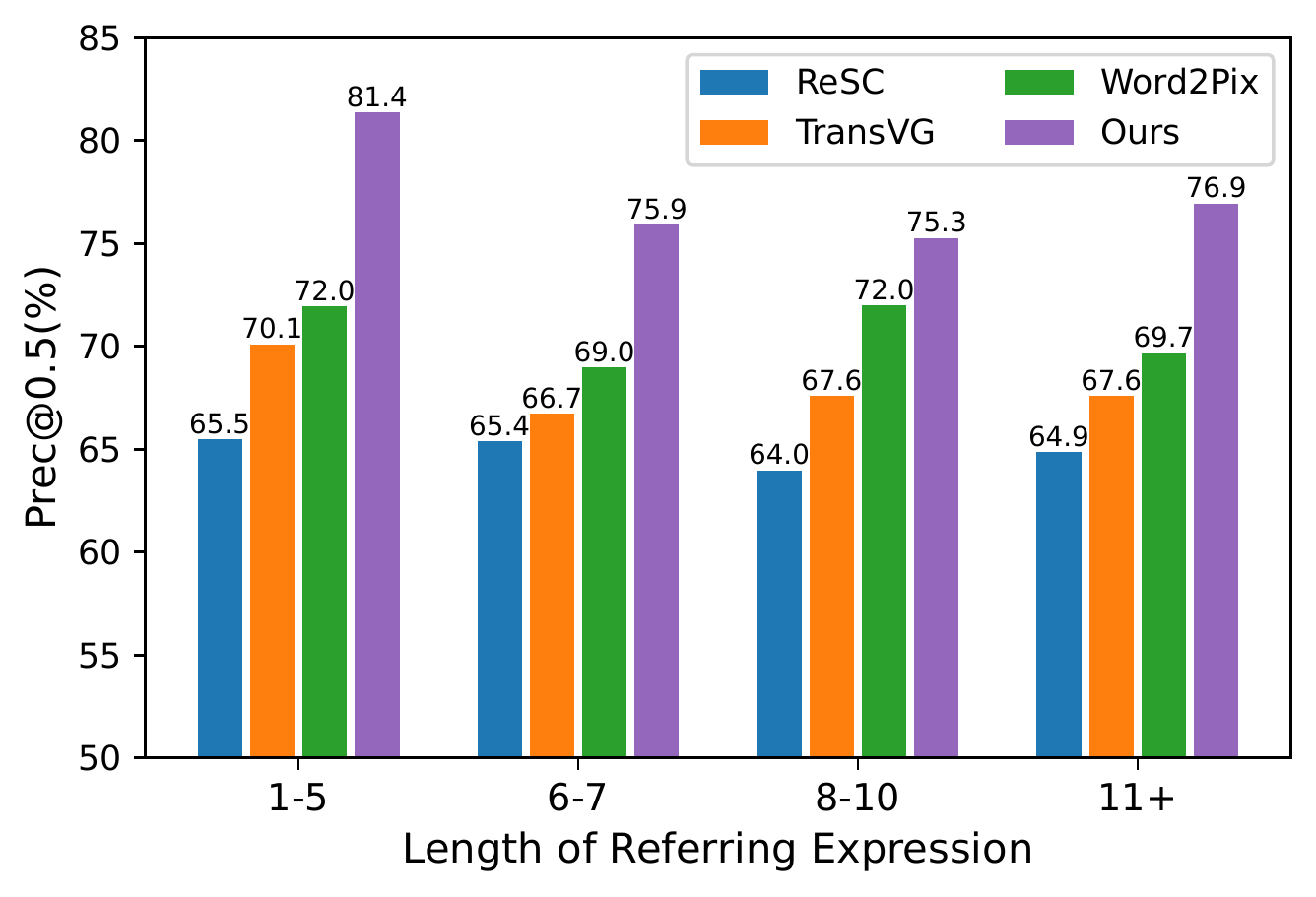}
    \caption{Comparison of accuracy under different lengths of referring expression on RefCOCOg-test. ReSC \cite{resc}, TransVG \cite{transvg}, Word2Pix \cite{word2pix}, and the proposed VG-LAW are compared.}
    \label{fig:acc_vs_length}
\end{figure}

\paragraph{Inference.}
At inference time, the input image is resized to $448 \times 448$, and the maximum length of referring expressions is set to 40.
Following the previous method \cite{mcn}, We set the threshold to 0.35 to realize the binarization of the RES prediction.
Without any post-processing operation, our framework directly outputs bounding boxes and segmentation maps specified by referring expressions.

\subsection{Comparisons with State-of-the-art Methods}
\label{experiments:sec3}
To estimate the effectiveness of the proposed VG-LAW framework, we conduct quantitative experiments on four widely used datasets, \textit{i.e.,} RefCOCO \cite{refcoco}, RefCOCO+ \cite{refcoco}, RefCOCOg \cite{refcocog}, and ReferItGame \cite{referitgame}.

\paragraph{REC Task.}
For the REC task, we compare the performance with state-of-the-art REC methods, including the two-stage methods \cite{mattnet,rvg-tree,cm-att-erase,ref-nms}, one-stage methods \cite{yang2019fast,resc,mcn,realgin,plv}, and transformer-based methods \cite{transvg,rt,seqtr,word2pix,yoro,qrnet}.
The main results are summarized to \cref{tab:rec_sota}.
It can be observed that VG-LAW achieves a significant performance improvement compared to the state-of-the-art two-stage method Ref-NMS \cite{ref-nms} and one-stage method PLV-FPN \cite{plv}.
When comparing to the transformer-based method QRNet \cite{qrnet}, which modified the visual backbone by inserting language-aware spatial and channel attention modules, our method has better performance with +2.62\%/ +3.47\%/ +0.82\% on RefCOCO, +3.43\%/ +4.87\%/ +3.69\% on RefCOCO+, +5.01\%/ +3.93\% on RefCOCOg, and +2.61\% on ReferItGame. 
QRNet \cite{qrnet} follows the TransVG \cite{transvg} framework, both of which use the transformer encoder-based cross-modal interaction module.
Compared to them, VG-LAW achieves better performance without complex cross-modal interaction modules.
Furthermore, our method significantly outperforms MCN \cite{mcn} and RefTR \cite{rt} based on joint training of REC and RES.

\paragraph{RES Task.}
For the RES task, we compare the performance with state-of-the-art methods \cite{cgan,mcn,lts,vlt,rt,seqtr,lavt}, and the main results are summarized to \cref{tab:res_sota}.
Compared with state-of-the-art RES method LAVT \cite{lavt}, VG-LAW achieves better \emph{mIoU} with +1.16\%/ +0.62\%/ +1.95\% on RefCOCO, +2.01\%/ +2.42\% on RefCOCOg, and comparable \emph{mIoU} with +0.82\%/ -0.59\%/ -0.34\% on RefCOCO+.
When comparing the models trained with or without multi-task settings, it can also be observed that consistent performance gains are achieved across all the datasets and splits.
As REC can provide localization information of the referred object, such coarse-grained supervision can slightly improve the segmentation accuracy in RES.

\paragraph{Analysis of Referring Expression Length.}
As the visual backbone in VG-LAW extracts features purely perceptually, it is of concern whether it can handle long and complex referring expressions.
ReSC \cite{resc} reveals that one-stage methods may ignore detailed descriptions in complex referring expressions and lead to poor performance.
Following that, we evaluate the REC performance on referring expressions of different lengths, as illustrated in \cref{fig:acc_vs_length}.
VG-LAW performs better than ReSC, TransVG \cite{transvg} and Word2Pix \cite{word2pix}, with no significant performance degradation when the length of referring expressions varies from 6-7 to 11+.

\begin{table}[t]
  \centering
    \begin{tabular}{ccc|c}
    \toprule
    LAWG   & LAP   & MTH   & \emph{Prec@0.5}(\%) \\
    \midrule
    \Checkmark     &       &       & 74.89  \\
          & \Checkmark     &       & 74.37  \\
    \Checkmark     & \Checkmark     &       & 76.60  \\
    \Checkmark     & \Checkmark     & \Checkmark     & 77.22  \\
    \bottomrule
    \end{tabular}%
    \caption{Ablation experiments on ReferItGame \cite{referitgame} to evaluate the proposed language adaptive weight generation (LAWG), language adaptive pooling (LAP), and multi-task head (MTH).}
  \label{tab:ablation}%
\end{table}%

\begin{figure*}[t]
    \centering
    \includegraphics[width=0.98\textwidth]{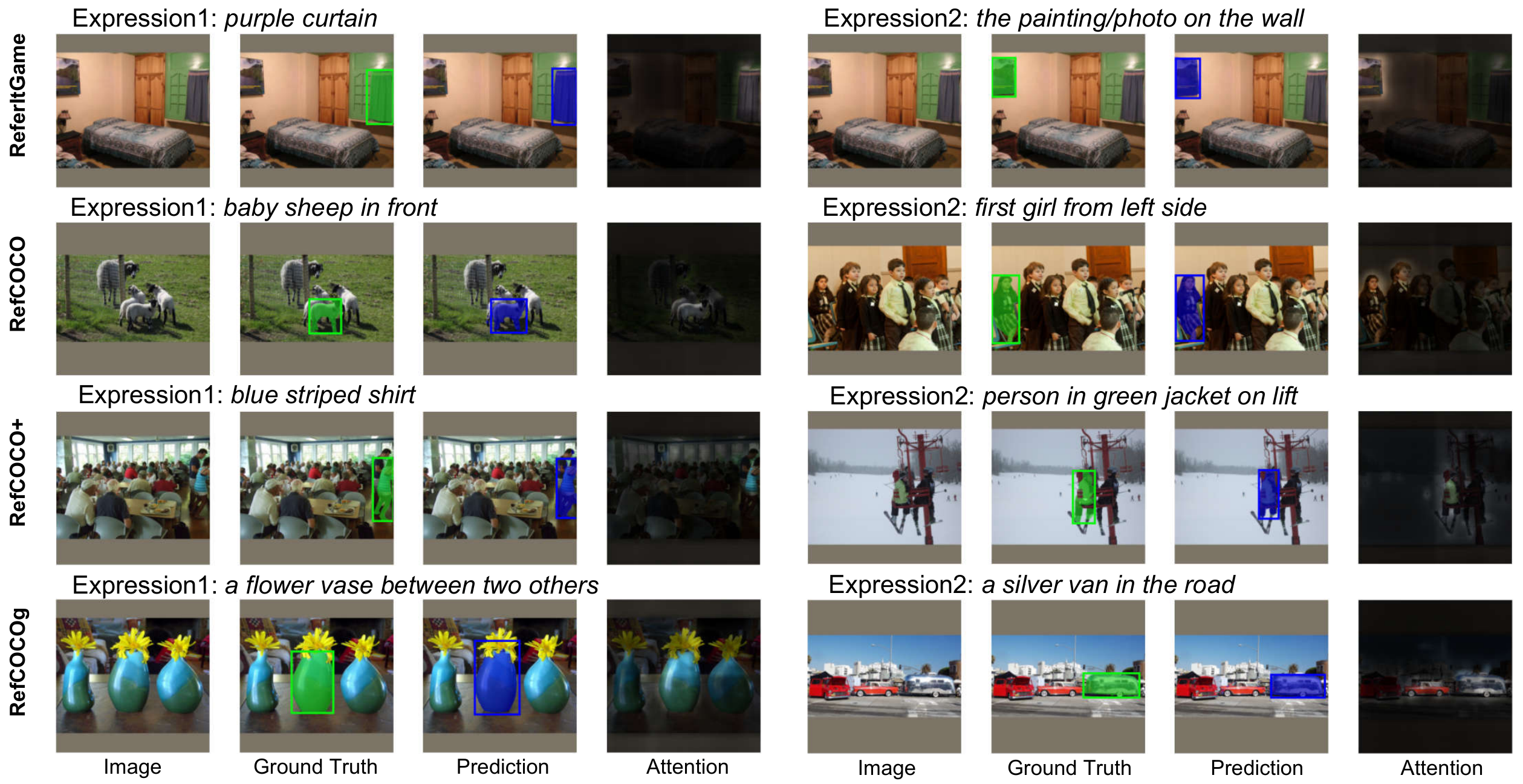}
    \caption{Qualitative results on the RefCOCO \cite{refcoco}, RefCOCO+ \cite{refcoco}, RefCOCOg \cite{refcocog}, and ReferItGame \cite{referitgame} datasets. Each dataset shows two examples. From left to right: the input image, the ground truth of REC and RES, the prediction of VG-LAW, and the attention of the visual backbone with language-adaptive weights.}
    \label{fig:qualitative}
\end{figure*}

\begin{figure}[t]
    \centering
    \includegraphics[width=0.45\textwidth]{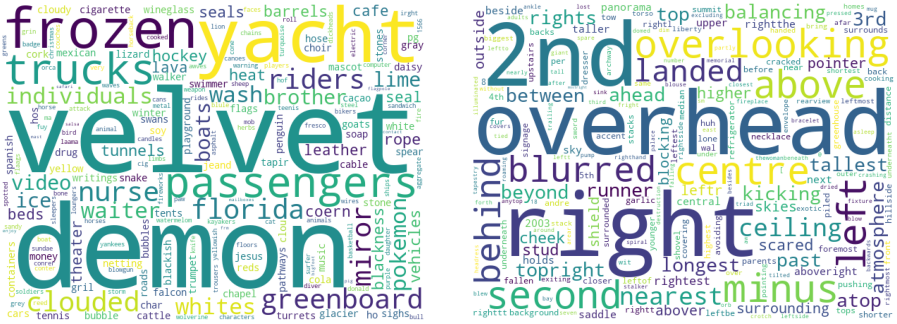}
    \caption{Wordcloud visualization of words assigned to the first and second halves of the visual backbone.}
    \label{fig:wordcloud}
\end{figure}

\subsection{Ablation Analysis}
\label{experiments:sec4}
To validate the effectiveness of our proposed modules, \textit{i.e.} language-adaptive weight generation, language-adaptive pooling, and multi-task head, we conduct ablation experiments on the REC dataset of ReferItGame, which is summarized in \cref{tab:ablation}.
When only using the LAWG, the visual features are pooled with global average pooling, and when only using the LAP, the visual backbone has fixed architecture and weights.
When only using the LAWG or the LAP, it can be observed that the model already achieves 74.89\% and 74.37\%, respectively, which is close to the 74.61\% reported by QRNet \cite{qrnet}.
When combined with the LAWG and LAP, further improvements can be brought by LAWG and LAP with +2.23\% and +1.71\%, respectively.
Benefiting from the auxiliary supervision of RES, our model equipped with the multi-task head can localize the referred objects better and achieve 77.22\%.

\subsection{Qualitative Results}
\label{experiments:sec5}

The qualitative results of the four datasets are shown in \cref{fig:qualitative}. 
It can be observed that our model can successfully locate and segment the referred objects, and the attention of the visual backbone can focus on the most relevant image regions, demonstrating the effectiveness of using language adaptive weights.
Taking the results on ReferItGame as an example, the visual backbone can dynamically filter out irrelevant regions for different expressions. For instance, when the ``purple curtain" is referred to, the regions related to the ``the painting/photo on the wall" are ignored.

In addition, we count the scores of words assigned to the first and second halves of the visual backbone, as shown in \cref{fig:wordcloud}. 
The scores are calculated by averaging attention score $\alpha_i$ in \cref{equ:token_attn} for each word, followed by softmax normalization along the layer dimension.
It can be observed that the shallow layers tend to the words describing individuals, such as the categories ``velvet" and ``yacht", and the deep layers tend to the words about contexts, such as the ordinal number ``2nd" and the position ``right".
\section{Conclusions and Liminations}

In this paper, we propose an active perception framework VG-LAW for visual grounding, based on the language adaptive weights. 
VG-LAW can directly inject the information of referring expressions into the weights of the visual backbone without modifying its architecture. 
Equipped with the proposed neat yet efficient multi-task head, VG-LAW achieves state-of-the-art performance for REC and RES tasks on widely used datasets.
The limitations of our method are two-fold: 
(1) VG-LAW is weak in interpretability, and the entire reasoning process is implicit, which makes it difficult to understand how the reasoning process works, and 
(2) the multi-task head predicts one instance at a time, which limits its application in phrase grounding.

\section{Acknowledgments}
This work is supported in part by National Key Research and Development Program of China under Grant 2020AAA0107400, National Natural Science Foundation of China under Grant U20A20222, National Science Foundation for Distinguished Young Scholars under Grant 62225605, Ant Group through CCF-Ant Research Fund, and sponsored by CCF-AFSG Research Fund, CAAI-HUAWEI MindSpore Open Fund, CCF-Zhipu AI Large Model Fund(CCF-Zhipu202302) as well as Hikvision Cooperation Fund.

{\small
\bibliographystyle{ieee_fullname}
\bibliography{egbib}

\begin{thebibliography}{10}\itemsep=-1pt

\bibitem{detr}
Nicolas Carion, Francisco Massa, Gabriel Synnaeve, Nicolas Usunier, Alexander
  Kirillov, and Sergey Zagoruyko.
\newblock End-to-end object detection with transformers.
\newblock In {\em ECCV}, pages 213--229, 2020.

\bibitem{ref-nms}
Long Chen, Wenbo Ma, Jun Xiao, Hanwang Zhang, and Shih-Fu Chang.
\newblock Ref-nms: Breaking proposal bottlenecks in two-stage referring
  expression grounding.
\newblock In {\em AAAI}, volume~35, pages 1036--1044, 2021.

\bibitem{chen2020dynamicconv}
Yinpeng Chen, Xiyang Dai, Mengchen Liu, Dongdong Chen, Lu Yuan, and Zicheng
  Liu.
\newblock Dynamic convolution: Attention over convolution kernels.
\newblock In {\em CVPR}, pages 11030--11039, 2020.

\bibitem{transvg}
Jiajun Deng, Zhengyuan Yang, Tianlang Chen, Wengang Zhou, and Houqiang Li.
\newblock Transvg: End-to-end visual grounding with transformers.
\newblock In {\em ICCV}, pages 1769--1779, 2021.

\bibitem{bert}
Jacob Devlin, Ming-Wei Chang, Kenton Lee, and Kristina Toutanova.
\newblock Bert: Pre-training of deep bidirectional transformers for language
  understanding.
\newblock {\em arXiv preprint arXiv:1810.04805}, 2018.

\bibitem{vlt}
Henghui Ding, Chang Liu, Suchen Wang, and Xudong Jiang.
\newblock Vision-language transformer and query generation for referring
  segmentation.
\newblock In {\em CVPR}, pages 16321--16330, 2021.

\bibitem{vit}
Alexey Dosovitskiy, Lucas Beyer, Alexander Kolesnikov, Dirk Weissenborn,
  Xiaohua Zhai, Thomas Unterthiner, Mostafa Dehghani, Matthias Minderer, Georg
  Heigold, Sylvain Gelly, et~al.
\newblock An image is worth 16x16 words: Transformers for image recognition at
  scale.
\newblock {\em arXiv preprint arXiv:2010.11929}, 2020.

\bibitem{saiapr12}
Hugo~Jair Escalante, Carlos~A Hern{\'a}ndez, Jesus~A Gonzalez, Aurelio
  L{\'o}pez-L{\'o}pez, Manuel Montes, Eduardo~F Morales, L~Enrique Sucar, Luis
  Villasenor, and Michael Grubinger.
\newblock The segmented and annotated iapr tc-12 benchmark.
\newblock {\em Computer vision and image understanding}, 114(4):419--428, 2010.

\bibitem{efn}
Guang Feng, Zhiwei Hu, Lihe Zhang, and Huchuan Lu.
\newblock Encoder fusion network with co-attention embedding for referring
  image segmentation.
\newblock In {\em CVPR}, pages 15506--15515, 2021.

\bibitem{ha2016hypernetworks}
David Ha, Andrew Dai, and Quoc~V Le.
\newblock Hypernetworks.
\newblock {\em arXiv preprint arXiv:1609.09106}, 2016.

\bibitem{he2017mask}
Kaiming He, Georgia Gkioxari, Piotr Doll{\'a}r, and Ross Girshick.
\newblock Mask r-cnn.
\newblock In {\em ICCV}, pages 2961--2969, 2017.

\bibitem{yoro}
Chih-Hui Ho, Srikar Appalaraju, Bhavan Jasani, R Manmatha, and Nuno
  Vasconcelos.
\newblock Yoro-lightweight end to end visual grounding.
\newblock In {\em ECCV2022 Workshops}, pages 3--23. Springer, 2023.

\bibitem{rvg-tree}
Richang Hong, Daqing Liu, Xiaoyu Mo, Xiangnan He, and Hanwang Zhang.
\newblock Learning to compose and reason with language tree structures for
  visual grounding.
\newblock {\em IEEE TPAMI}, 2019.

\bibitem{hu2016segmentation}
Ronghang Hu, Marcus Rohrbach, and Trevor Darrell.
\newblock Segmentation from natural language expressions.
\newblock In {\em ECCV}, pages 108--124, 2016.

\bibitem{cmpc}
Shaofei Huang, Tianrui Hui, Si Liu, Guanbin Li, Yunchao Wei, Jizhong Han, Luoqi
  Liu, and Bo Li.
\newblock Referring image segmentation via cross-modal progressive
  comprehension.
\newblock In {\em CVPR}, pages 10488--10497, 2020.

\bibitem{jia2016dynamicfilternet}
Xu Jia, Bert De~Brabandere, Tinne Tuytelaars, and Luc~V Gool.
\newblock Dynamic filter networks.
\newblock {\em NeurIPS}, 29, 2016.

\bibitem{lts}
Ya Jing, Tao Kong, Wei Wang, Liang Wang, Lei Li, and Tieniu Tan.
\newblock Locate then segment: A strong pipeline for referring image
  segmentation.
\newblock In {\em CVPR}, pages 9858--9867, 2021.

\bibitem{mdetr}
Aishwarya Kamath, Mannat Singh, Yann LeCun, Gabriel Synnaeve, Ishan Misra, and
  Nicolas Carion.
\newblock Mdetr-modulated detection for end-to-end multi-modal understanding.
\newblock In {\em ICCV}, pages 1780--1790, 2021.

\bibitem{referitgame}
Sahar Kazemzadeh, Vicente Ordonez, Mark Matten, and Tamara Berg.
\newblock Referitgame: Referring to objects in photographs of natural scenes.
\newblock In {\em EMNLP}, pages 787--798, 2014.

\bibitem{restr}
Namyup Kim, Dongwon Kim, Cuiling Lan, Wenjun Zeng, and Suha Kwak.
\newblock Restr: Convolution-free referring image segmentation using
  transformers.
\newblock In {\em CVPR}, pages 18145--18154, 2022.

\bibitem{imagenet}
Alex Krizhevsky, Ilya Sutskever, and Geoffrey~E Hinton.
\newblock Imagenet classification with deep convolutional neural networks.
\newblock In {\em NeurIPS}, 2012.

\bibitem{li2022omni}
Chao Li, Aojun Zhou, and Anbang Yao.
\newblock Omni-dimensional dynamic convolution.
\newblock {\em arXiv preprint arXiv:2209.07947}, 2022.

\bibitem{rt}
Muchen Li and Leonid Sigal.
\newblock Referring transformer: A one-step approach to multi-task visual
  grounding.
\newblock {\em NeurIPS}, 34:19652--19664, 2021.

\bibitem{li2020dcd}
Yunsheng Li, Yinpeng Chen, Xiyang Dai, Dongdong Chen, Ye Yu, Lu Yuan, Zicheng
  Liu, Mei Chen, Nuno Vasconcelos, et~al.
\newblock Revisiting dynamic convolution via matrix decomposition.
\newblock In {\em ICLR}, 2020.

\bibitem{li2022vitdet}
Yanghao Li, Hanzi Mao, Ross Girshick, and Kaiming He.
\newblock Exploring plain vision transformer backbones for object detection.
\newblock {\em arXiv preprint arXiv:2203.16527}, 2022.

\bibitem{plv}
Yue Liao, Aixi Zhang, Zhiyuan Chen, Tianrui Hui, and Si Liu.
\newblock Progressive language-customized visual feature learning for one-stage
  visual grounding.
\newblock {\em IEEE TIP}, 31:4266--4277, 2022.

\bibitem{focal}
Tsung-Yi Lin, Priya Goyal, Ross Girshick, Kaiming He, and Piotr Doll{\'a}r.
\newblock Focal loss for dense object detection.
\newblock In {\em ICCV}, pages 2980--2988, 2017.

\bibitem{mscoco}
Tsung-Yi Lin, Michael Maire, Serge Belongie, James Hays, Pietro Perona, Deva
  Ramanan, Piotr Doll{\'a}r, and C~Lawrence Zitnick.
\newblock Microsoft coco: Common objects in context.
\newblock In {\em ECCV}, pages 740--755, 2014.

\bibitem{liu2019learning}
Daqing Liu, Hanwang Zhang, Feng Wu, and Zheng-Jun Zha.
\newblock Learning to assemble neural module tree networks for visual
  grounding.
\newblock In {\em ICCV}, pages 4673--4682, 2019.

\bibitem{cm-att-erase}
Xihui Liu, Zihao Wang, Jing Shao, Xiaogang Wang, and Hongsheng Li.
\newblock Improving referring expression grounding with cross-modal
  attention-guided erasing.
\newblock In {\em CVPR}, pages 1950--1959, 2019.

\bibitem{adamw}
Ilya Loshchilov and Frank Hutter.
\newblock Decoupled weight decay regularization.
\newblock {\em arXiv preprint arXiv:1711.05101}, 2017.

\bibitem{cgan}
Gen Luo, Yiyi Zhou, Rongrong Ji, Xiaoshuai Sun, Jinsong Su, Chia-Wen Lin, and
  Qi Tian.
\newblock Cascade grouped attention network for referring expression
  segmentation.
\newblock In {\em ACM MM}, pages 1274--1282, 2020.

\bibitem{mcn}
Gen Luo, Yiyi Zhou, Xiaoshuai Sun, Liujuan Cao, Chenglin Wu, Cheng Deng, and
  Rongrong Ji.
\newblock Multi-task collaborative network for joint referring expression
  comprehension and segmentation.
\newblock In {\em CVPR}, pages 10034--10043, 2020.

\bibitem{refcocog}
Junhua Mao, Jonathan Huang, Alexander Toshev, Oana Camburu, Alan~L Yuille, and
  Kevin Murphy.
\newblock Generation and comprehension of unambiguous object descriptions.
\newblock In {\em CVPR}, pages 11--20, 2016.

\bibitem{dice}
Fausto Milletari, Nassir Navab, and Seyed-Ahmad Ahmadi.
\newblock V-net: Fully convolutional neural networks for volumetric medical
  image segmentation.
\newblock In {\em 2016 fourth international conference on 3D vision}, pages
  565--571. IEEE, 2016.

\bibitem{refcocog-umd}
Varun~K Nagaraja, Vlad~I Morariu, and Larry~S Davis.
\newblock Modeling context between objects for referring expression
  understanding.
\newblock In {\em ECCV}, pages 792--807, 2016.

\bibitem{redmon2018yolov3}
Joseph Redmon and Ali Farhadi.
\newblock Yolov3: An incremental improvement.
\newblock {\em arXiv preprint arXiv:1804.02767}, 2018.

\bibitem{faster}
Shaoqing Ren, Kaiming He, Ross Girshick, and Jian Sun.
\newblock Faster r-cnn: Towards real-time object detection with region proposal
  networks.
\newblock {\em NeurIPS}, 28, 2015.

\bibitem{giou}
Hamid Rezatofighi, Nathan Tsoi, JunYoung Gwak, Amir Sadeghian, Ian Reid, and
  Silvio Savarese.
\newblock Generalized intersection over union: A metric and a loss for bounding
  box regression.
\newblock In {\em CVPR}, pages 658--666, 2019.

\bibitem{transformer}
Ashish Vaswani, Noam Shazeer, Niki Parmar, Jakob Uszkoreit, Llion Jones,
  Aidan~N Gomez, {\L}ukasz Kaiser, and Illia Polosukhin.
\newblock Attention is all you need.
\newblock {\em NeurIPS}, 30, 2017.

\bibitem{yang2019condconv}
Brandon Yang, Gabriel Bender, Quoc~V Le, and Jiquan Ngiam.
\newblock Condconv: Conditionally parameterized convolutions for efficient
  inference.
\newblock {\em NeurIPS}, 32, 2019.

\bibitem{resc}
Zhengyuan Yang, Tianlang Chen, Liwei Wang, and Jiebo Luo.
\newblock Improving one-stage visual grounding by recursive sub-query
  construction.
\newblock In {\em ECCV}, pages 387--404, 2020.

\bibitem{yang2019fast}
Zhengyuan Yang, Boqing Gong, Liwei Wang, Wenbing Huang, Dong Yu, and Jiebo Luo.
\newblock A fast and accurate one-stage approach to visual grounding.
\newblock In {\em ICCV}, pages 4683--4693, 2019.

\bibitem{lavt}
Zhao Yang, Jiaqi Wang, Yansong Tang, Kai Chen, Hengshuang Zhao, and Philip~HS
  Torr.
\newblock Lavt: Language-aware vision transformer for referring image
  segmentation.
\newblock In {\em CVPR}, pages 18155--18165, 2022.

\bibitem{qrnet}
Jiabo Ye, Junfeng Tian, Ming Yan, Xiaoshan Yang, Xuwu Wang, Ji Zhang, Liang He,
  and Xin Lin.
\newblock Shifting more attention to visual backbone: Query-modulated
  refinement networks for end-to-end visual grounding.
\newblock In {\em CVPR}, pages 15502--15512, 2022.

\bibitem{mattnet}
Licheng Yu, Zhe Lin, Xiaohui Shen, Jimei Yang, Xin Lu, Mohit Bansal, and
  Tamara~L Berg.
\newblock Mattnet: Modular attention network for referring expression
  comprehension.
\newblock In {\em CVPR}, pages 1307--1315, 2018.

\bibitem{refcoco}
Licheng Yu, Patrick Poirson, Shan Yang, Alexander~C Berg, and Tamara~L Berg.
\newblock Modeling context in referring expressions.
\newblock In {\em ECCV}, pages 69--85, 2016.

\bibitem{word2pix}
Heng Zhao, Joey~Tianyi Zhou, and Yew-Soon Ong.
\newblock Word2pix: Word to pixel cross-attention transformer in visual
  grounding.
\newblock {\em IEEE Trans. Neural Networks and Learning Systems.}, 2022.

\bibitem{realgin}
Yiyi Zhou, Rongrong Ji, Gen Luo, Xiaoshuai Sun, Jinsong Su, Xinghao Ding,
  Chia-Wen Lin, and Qi Tian.
\newblock A real-time global inference network for one-stage referring
  expression comprehension.
\newblock {\em IEEE Trans. Neural Networks and Learning Systems.}, 2021.

\bibitem{seqtr}
Chaoyang Zhu, Yiyi Zhou, Yunhang Shen, Gen Luo, Xingjia Pan, Mingbao Lin, Chao
  Chen, Liujuan Cao, Xiaoshuai Sun, and Rongrong Ji.
\newblock Seqtr: A simple yet universal network for visual grounding.
\newblock {\em arXiv preprint arXiv:2203.16265}, 2022.

\end{thebibliography}
}

\end{document}